\documentclass{sig-alternate-2013}

\usepackage{epsfig}
\usepackage{graphicx}
\usepackage{amsmath}
\usepackage{amssymb}

\usepackage[breaklinks=true,bookmarks=false]{hyperref}

\newfont{\mycrnotice}{ptmr8t at 7pt}
\newfont{\myconfname}{ptmri8t at 7pt}
\let\crnotice\mycrnotice%
\let\confname\myconfname%

\permission{Permission to make digital or hard copies of all or part of this work for personal or classroom use is granted without fee provided that copies are not made or distributed for profit or commercial advantage and that copies bear this notice and the full citation on the first page. Copyrights for components of this work owned by others than the author(s) must be honored. Abstracting with credit is permitted. To copy otherwise, or republish, to post on servers or to redistribute to lists, requires prior specific permission and/or a fee. Request permissions from Permissions@acm.org.}
\conferenceinfo{KDD'15,}{August 10-13, 2015, Sydney, NSW, Australia. \\ 
	{\mycrnotice{Copyright is held by the owner/author(s). Publication rights licensed to ACM.}}}
\copyrightetc{ACM \the\acmcopyr}
\crdata{978-1-4503-3664-2/15/08\ ...\$15.00.\\
	DOI: http://dx.doi.org/10.1145/2783258.2788621 }

\begin{document}



\title{Visual Search at Pinterest}


\numberofauthors{1} 
\author{\alignauthor Yushi Jing$^{^1}$, David Liu$^{^1}$, Dmitry Kislyuk$^{^1}$, Andrew Zhai$^{^1}$, Jiajing Xu$^{^1}$, Jeff Donahue$^{^{1,2}}$, Sarah Tavel$^{^1}$\\
\affaddr{$^{^1}$Visual Discovery, Pinterest}\\
\affaddr{$^{^2}$University of California, Berkeley}\\
\email{\{jing, dliu, dkislyuk, andrew, jiajing, jdonahue, sarah\}@pinterest.com}
}

\maketitle


\begin{abstract}

We demonstrate that, with the availability of distributed computation platforms such as Amazon Web Services and open-source tools, it is possible for a small engineering team to build, launch and maintain a cost-effective, large-scale visual search system with widely available tools. We also demonstrate, through a comprehensive set of live experiments at Pinterest, that content recommendation powered by visual search improve user engagement. By sharing our implementation details and the experiences learned from launching a commercial visual search engines from scratch, we hope visual search are more widely incorporated into today's commercial applications.
\newline
\newline
\textbf{Please see an updated version of the paper, Visual Discovery at Pinterest, presented at World Wide Web (WWW) 2017.} 



\end{abstract}




\section{Introduction}

Visual search, or content-based image retrieval~\cite{Datta:2008}, is an active research area driven in part by the explosive growth of online photos and the popularity of search engines. Google Goggles, Google Similar Images and Amazon Flow are several examples of commercial visual search systems.  Although significant progress has been made in building Web-scale visual search systems, there are few publications describing end-to-end architectures deployed on commercial applications.  This is in part due to the complexity of real-world visual search systems, and in part due to business considerations to keep core search technology proprietary.

We faced two main challenges in deploying a commercial visual search system at Pinterest. First, as a startup we needed to control the development cost in the form of both human and computational resources.  For example, feature computation can become expensive with a large and continuously growing image collection, and with engineers constantly experimenting with new features to deploy, it is vital for our system to be both scalable and cost effective. 
Second, the success of a commercial application is measured by the benefit it brings to the users (e.g. improved user engagement) relative to the cost of development and maintenance. As a result, our development progress needs to be frequently validated through A/B experiments with live user traffic.

\begin{figure}
\centering \includegraphics[width=3in]{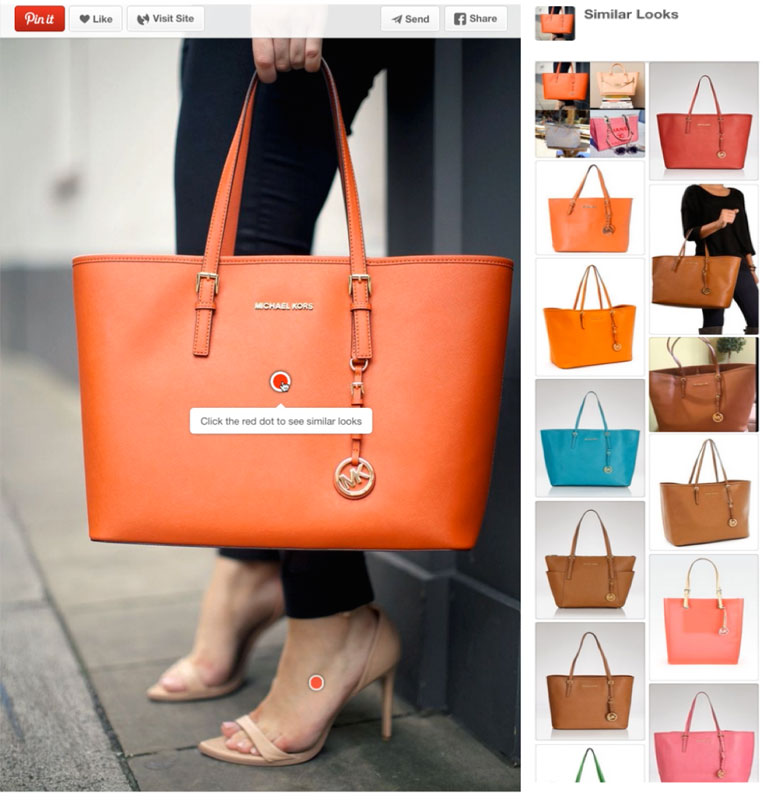}
\caption{Similar Looks:   We apply object detection to localize products such as bags and shoes. In this prototype, users click on objects of interest to view similar-looking products.}
\label{fig:similarlooks}
\end{figure}

In this paper, we describe our approach to deploy a commercial visual search system with those two challenges in mind. We makes two main contributions.

Our first contribution is to present our scalable and cost effective visual search implementation using widely available tools, feasible for a small engineering team to implement. Section 2.1 describes our simple and pragmatic approach to speeding up and improving the accuracy of object detection and localization that exploits the rich metadata available at Pinterest.  By decoupling the difficult (and computationally expensive) task of multi-class object detection into category classification followed by per-category object detection, we only need to run (expensive) object detectors on images with high probability of containing the object.  Section 2.2 presents our distributed pipeline to incrementally add or update image features using Amazon Web Services, which avoids wasteful re-computation of unchanged image features.  Section 2.3 presents our distributed indexing and search infrastructure built on top of widely available tools.

\begin{figure}
\centering \includegraphics[width=3 in]{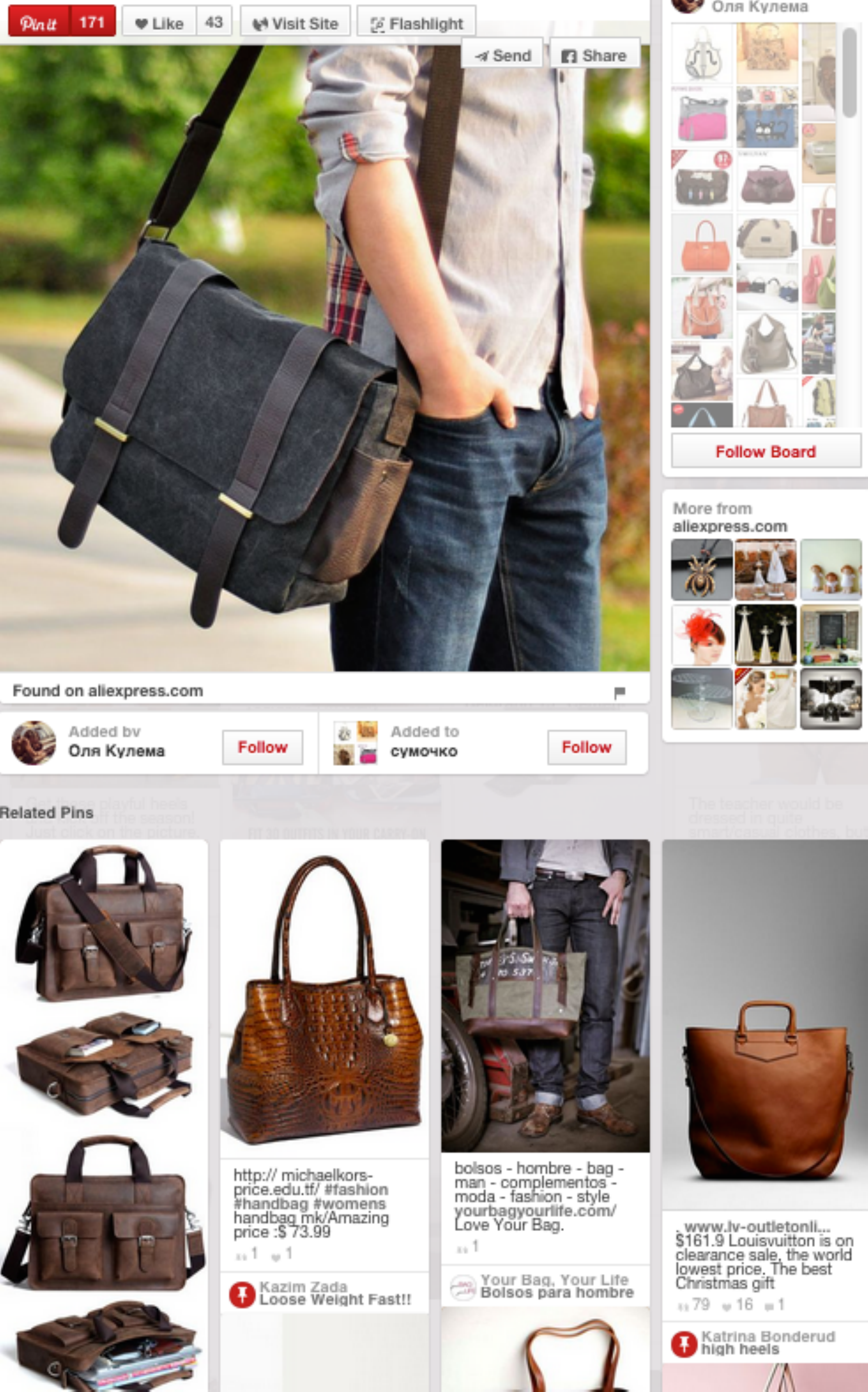}
\caption{Related Pins:  Pins are selected based on the curation graph.}
\label{fig:relatedpins}
\end{figure}

Our second contribution is to share results of deploying our visual search infrastructure in two product applications: \textit{Related Pins} (Section 3) and \textit{Similar Looks} (Section 4).
For each application, we use application-specific data sets to evaluate the effectiveness of each visual search component (object detection, feature representations for similarity) in isolation.
After deploying the end-to-end system, we use A/B tests to measure user engagement on live traffic.

Related Pins (Figure~\ref{fig:relatedpins}) is a feature that recommends Pins based on the Pin the user is currently viewing. These recommendations are primarily generated from the ``curation graph'' of users, boards, and Pins. However, there is a long tail of less popular Pins without recommendations. Using visual search, we generate recommendations for almost all Pins on Pinterest.
Our second application, \textit{Similar Looks} (Figure~\ref{fig:similarlooks} )  is a discovery experience we tested specifically for fashion Pins.  It allowed users to select a visual query from regions of interest (e.g. a bag or a pair of shoes) and identified visually similar Pins for users to explore or purchase.
Instead of using the whole image, visually similarity is computed between the localized objects in the query and database images.  To our knowledge, this is the first published work on object detection/localization in a commercially deployed visual search system.



Our experiments demonstrate that 1) one can achieve very low false positive rate (less than 1\%) with good detection rate by combining the object detection/localization methods with metadata, and 2) using feature representations from the VGG~\cite{Simonyan14c}~\cite{vggnet2014} model significantly improves visual search accuracy on our Pinterest benchmark datasets, and 3) we observe significant gains in user engagement when visual search is used to power Related Pins and Similar Looks applications.

\section{Visual Search Architecture at Pinterest}
\label{sec:arch}

Pinterest is a visual bookmarking tool that helps users discover and save creative ideas.  Users \textit{pin} images to \textit{boards}, which are curated collections around particular themes or topics.  This human-curated user-board-image graph contains a rich set of information about the images and their semantic relations to each other.   For example, when an image is pinned to a board, it is implies a ``curatorial link" between the new board and all other boards the image appears in.  Metadata, such as image annotations, can then be propagated through these links to form a rich description of the image, the image board and the users.

Since the image is the focus of each pin, visual features play a large role in finding interesting, inspiring and relevant content for users.  In this section we describe the end-to-end implementation of a visual search system that indexes billions of images on Pinterest. We address the challenges of developing a real-world visual search system that balances cost constraints with the need for fast prototyping. We describe 1) the features that we extract from images, 2) our infrastructure for distributed and incremental feature extraction, and 3) our real-time visual search service.

%
%

\subsection{Image Representation and Features}
\label{sec:features}

We extract a variety of features from images, including local features and ``deep features'' extracted from the activation of intermediate layers of a deep convolutional network.
%
%
The deep features come from convolutional neural networks (CNNs) based on the AlexNet~\cite{alexnet12} and VGG~\cite{Simonyan14c} architectures. We used the feature representations from \textit{fc6} and \textit{fc8} layers. These features are binarized for representation efficiency and compared using Hamming distance. We use open-source Caffe~\cite{jia2014caffe} to perform training and inference of our CNNs on multi-GPU machines.

The system also extracts \textit{salient color signatures} from images.
Salient colors are computed by first detecting salient regions~\cite{saliency13,saliency14} of the images and then applying $k$-means clustering to the Lab pixel values of the salient pixels.  Cluster centroids and weights are stored as the color signature of the image.

%

\subsubsection*{Two-step Object Detection and Localization}

One feature that is particularly relevant to Pinterest is the presence of certain object classes, such as bags, shoes, watches, dresses, and sunglasses. We adopted a two-step detection approach that leverages the abundance of weak text labels on Pinterest images. Since images are pinned many times onto many boards, aggregated pin descriptions and board titles provide a great deal of information about the image. A text processing pipeline within Pinterest extracts relevant annotations for images from the raw text, producing short phrases associated with each image.

We use these annotations to determine which object detectors to run. In Figure~\ref{fig:similarlooks}, we first determined that the image was likely to contain bags and shoes, and then proceeded to apply visual object detectors for those object classes. By first performing category classification, we only need to run the object detectors on images with a high prior likelihood of matching, reducing computational cost as well as false positives.


Our initial approach for object detection was a heavily optimized implementation of cascading deformable part-based models~\cite{partsbased10}. This detector outputs a bounding box for each detected object, from which we extract visual descriptors for the object.  Our recent efforts have focused on investigating the feasibility and performance of deep learning based object detectors~\cite{rcnn,spp,erhan} as a part of our two-step detection/localization pipeline.

Our experiment results in Section~\ref{sec:evaluationobject} show that our system achieved a very low false positive rate (less than 1\%), which was vital for our application. This two-step approach also enables us to incorporate other signals into the category classification.  The use of both text and visual signals for object detection and localization is widely used~\cite{Berg:2004}~\cite{bengio13}~\cite{visualrank} for Web image retrieval and categorization. 

\begin{figure}
\centering
\includegraphics[width=3.4 in]{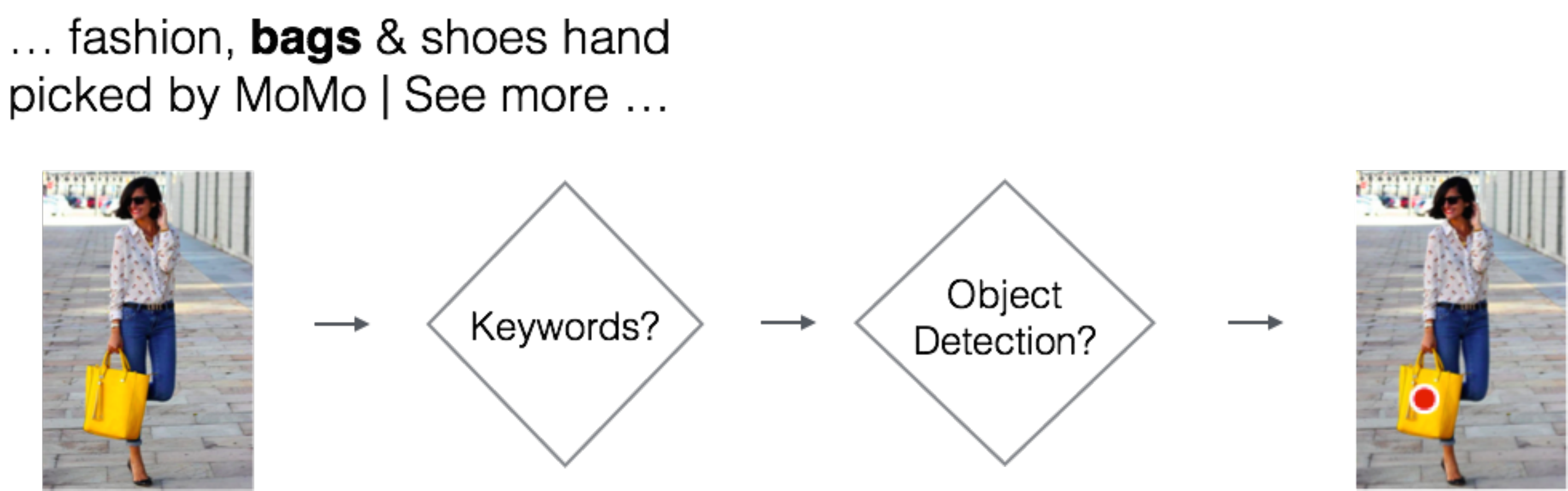}
\caption{Instead of running all object detectors on all images, we first predict the image categories using textual metadata, and then apply object detection modules specific to the predicted category.}
\label{fig:tiered}
\end{figure}

\subsubsection*{Click Prediction }
When users browse on Pinterest, they can interact with a pin by clicking to view it full screen (``close-up") and subsequently clicking through to the off-site source of the content (a click-through). For each image, we predict close-up rate (CUR) and click-through rate (CTR) based on its visual features. We trained a CNN to learn a mapping from images to the probability of a user bringing up the close-up view or clicking through to the content. Both  CUR and CTR are helpful for applications like search ranking, recommendation systems and ads targeting since we often need to know which images are more likely to get attention from users based on their visual content.

CNNs have recently become the dominant approach to many semantic prediction tasks involving visual inputs, including classification~\cite{LeCun:1989,alexnet12,googlenet2014,vggnet2014,imagenet2014,karpathy}, detection~\cite{rcnn,spp,erhan}, and segmentation~\cite{jonevan}. Training a full CNN to learn good representation can be time-consuming and requires a very large corpus of data. We apply transfer learning to our model by retaining the low-level visual representations from models trained for other computer vision tasks. The top-level layers of the network are fine-tuned for our specific task. This saves substantial training time and leverages the visual features learned from a much larger corpus than that of the target task. We use Caffe to perform this transfer learning.


\begin{figure}
	\centering
	\includegraphics[width=3.4in]{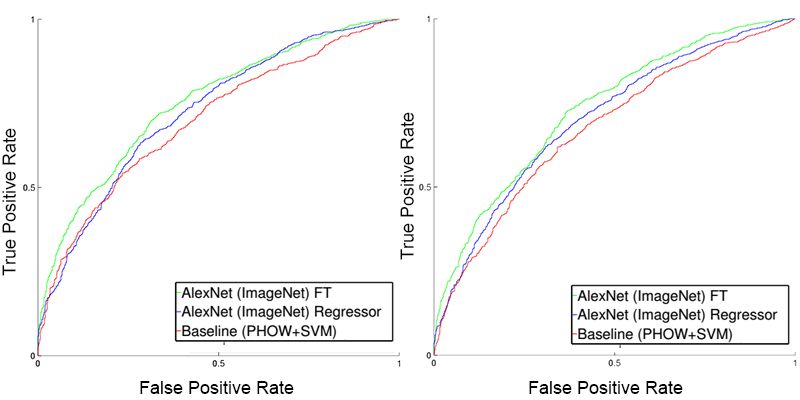}
	\caption{ROC curves for CUR prediction (left) and CTR prediction (right). }
	\label{fig:ads_ctr_cur}
\end{figure}

Figure~\ref{fig:ads_ctr_cur} depicts receiver operating characteristic (ROC) curves for our CNN-based method, compared with a baseline based on a ``traditional'' computer vision pipeline: a SVM trained with binary labels on a pyramid histogram of words (PHOW), which performs well on object recognition datasets such as Caltech-101.  Our CNN-based approach outperforms the PHOW-SVM baseline, and fine-tuning the CNN from end-to-end yields a significant performance boost as well.   A similar approach was also applied to the task of detecting pornographic images uploaded to Pinterest with good results~\footnote{By fine-tuning a network for three-class classification of \textit{ignore}, \textit{softcore}, and \textit{porn} images, we are able to achieve a validation accuracy of 83.2\%. When formulated as a binary classification between \textit{ignore} and \textit{softcore}/\textit{porn} categories, the classifier achieved an AUC score of 93.56\%.}.
\subsection{Incremental Fingerprinting Service}

\label{sec:incremental}

\label{sec:ifs}
Most of our vision applications depend on having a complete collection of image features, stored in a format amenable to bulk processing.
Keeping this data up-to-date is challenging; because
our collection comprises over a billion unique images, it is critical to update the feature set incrementally and avoid unnecessary re-computation whenever possible.



We built a system called the \textit{Incremental Fingerprinting Service}, which computes image features for all Pinterest images using a cluster of workers on Amazon EC2. It incrementally updates the collection of features under two main change scenarios: new images uploaded to Pinterest, and feature evolution (features added/modified by engineers).

\begin{figure}
\centering
\includegraphics[height=5.5 in]{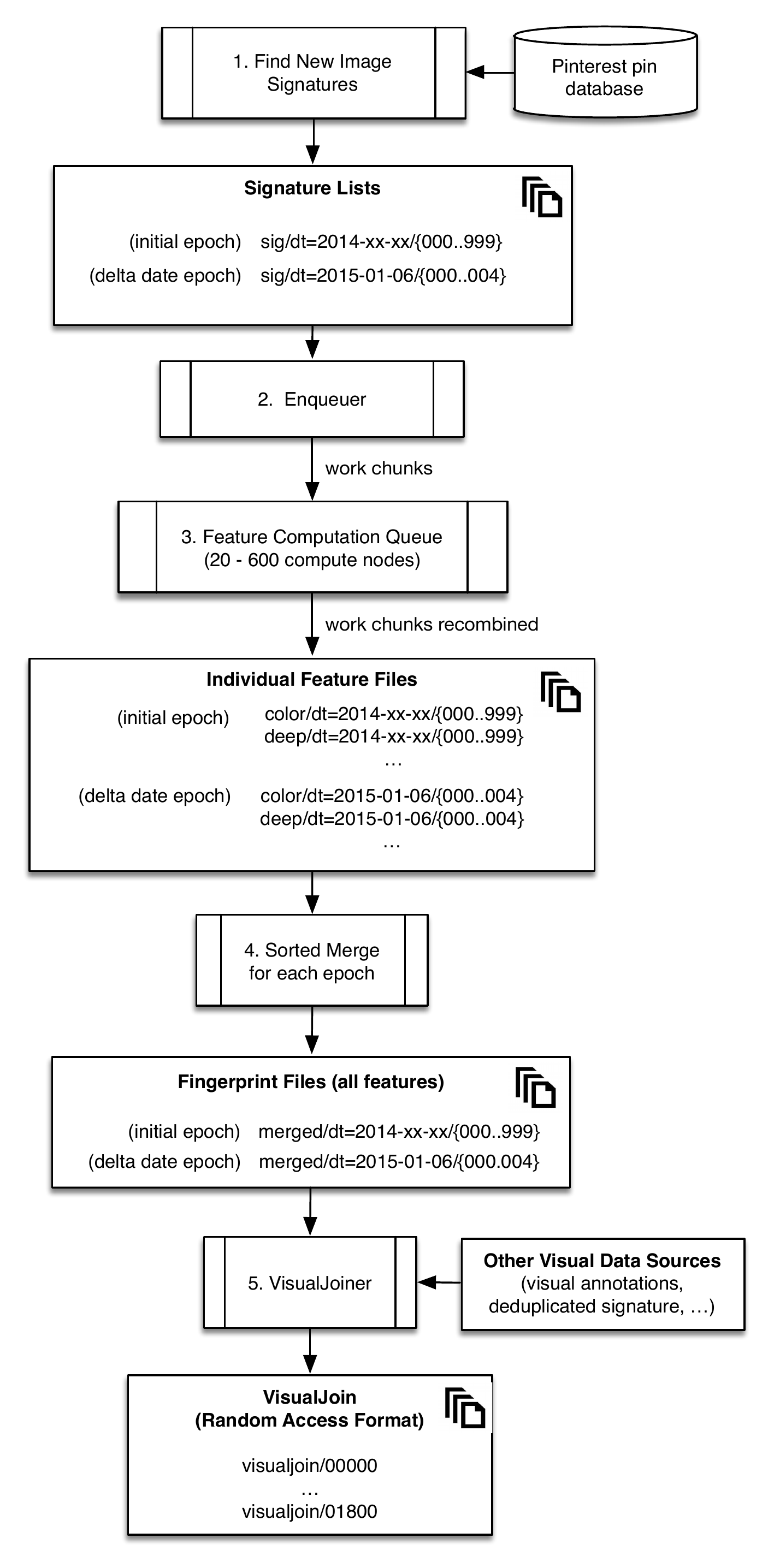}
\caption{Examples of outputs generated by incremental fingerprint update pipeline. The initial run is shown as 2014-xx-xx which includes all the images created before that run.}
\label{fig:ifu_2}
\end{figure}

Our approach is to split the image collection into \textit{epochs} grouped by upload date, and to maintain a separate feature store for each version of each feature type (global, local, deep features). Features are stored in bulk on Amazon S3, organized by feature type, version, and date. 
When the data is fully up-to-date, each feature store contains all the epochs. On each run, the system detects missing epochs for each feature and enqueues jobs into a distributed queue to populate those epochs.


This storage scheme enables incremental updates as follows. Every day, a new epoch is added to our collection with that day's unique uploads, and we generate the missing features for that date. Since old images do not change, their features are not recomputed.  If the algorithm or parameters for generating a feature are modified, or if a new feature is added, a new feature store is started and all of the epochs are computed for that feature.  Unchanged features are not affected.

We copy these features into various forms for more convenient access by other jobs: features are merged to form a \textit{fingerprint} containing all available features of an image, and fingerprints are copied into sharded, sorted files for random access by image signature (MD5 hash). These joined fingerprint files are regularly re-materialized, but the expensive feature computation needs only be done once per image.

A flow chart of the incremental fingerprint update process is shown in Figure~\ref{fig:ifu_2}. It consists of five main jobs:  job (1) compiles a list of newly uploaded image signatures and groups them by date into epochs. We randomly divide each epoch into sorted shards of approximately 200,000 images to limit the size of the final fingerprint files.  Job (2) identifies missing epochs in each feature store and enqueues jobs into PinLater (a distributed queue service similar to Amazon SQS). The jobs subdivide the shards into ``work chunks", tuned such that each chunk takes approximate 30 minutes to compute.
Job (3) runs on an automatically-launched cluster of EC2 instances, scaled depending on the size of the update. Spot instances can be used; if an instance is terminated, its job is rescheduled on another worker. The output of each work chunk is saved onto S3, and eventually recombined into feature files corresponding to the original shards.

Job (4) merges the individual feature shards into a unified fingerprint containing all of the available features for each image. Job (5) merges all the epochs into a sorted, sharded HFile format allowing for random access.

The initial computation of all available features on all images, takes a little over a day using a cluster of several hundred 32-core machines, and produces roughly 5~TB of feature data.  The steady-state requirement to process new images incrementally is only about 5 machines.

\subsection{Search Infrastructure}
\begin{figure}
	\centering
	\includegraphics[width=3.2in]{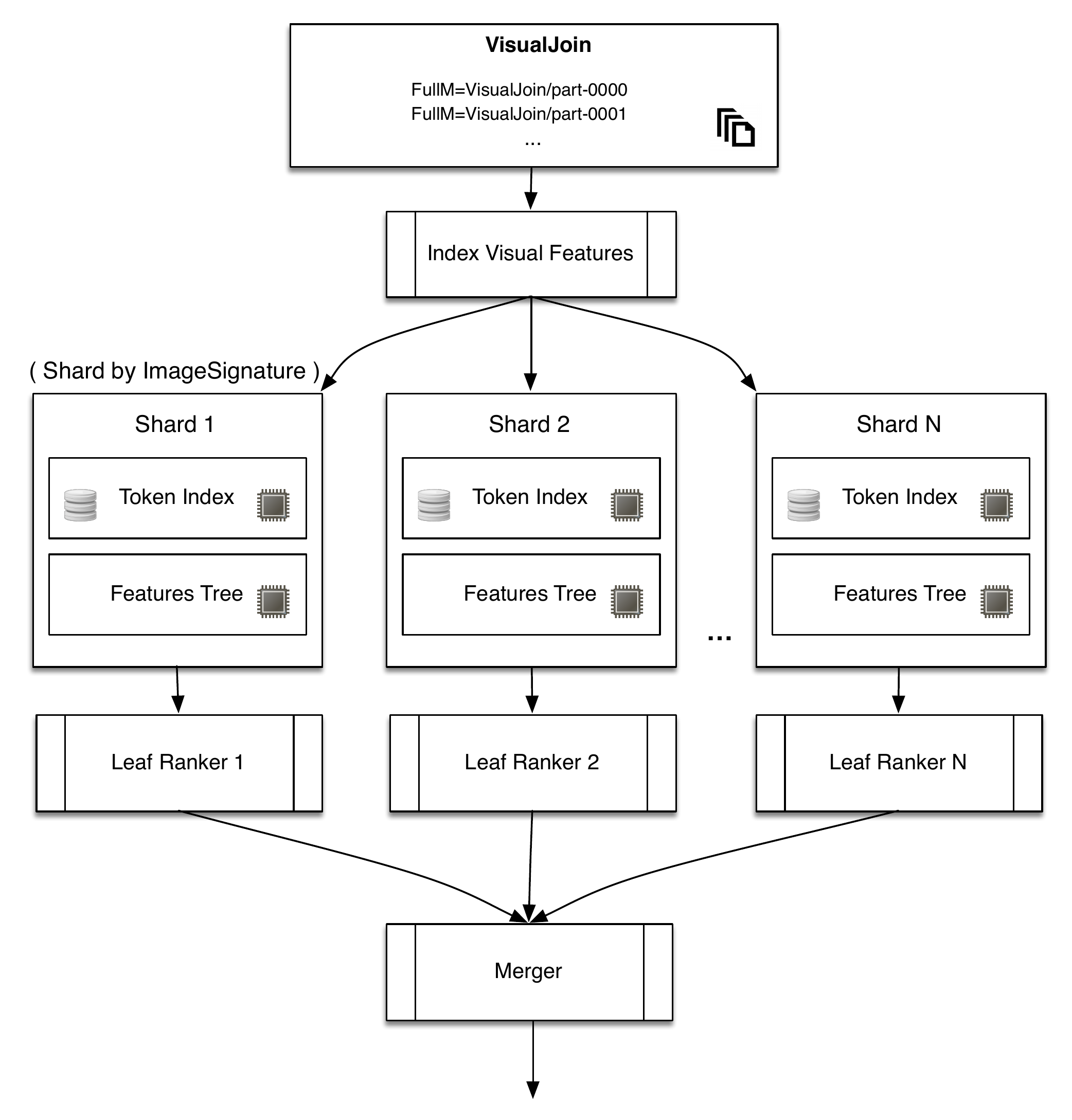}
	\caption{A flow chart of the distributed visual search pipeline.}
	\label{fig:search_pipeline}
\end{figure}

At Pinterest, there are several use cases for a distributed visual search system.  One use case is to explore similar looking products (Pinterest Similar Looks), and others include near-duplicate detection and content recommendation. In all these applications, visually similar results are computed from distributed indices built on top of the \textit{visualjoins} generated in the previous section. Since each use case has a different set of performance and cost requirements, our infrastructure is designed to be flexible and re-configurable.  A flow chart of the search infrastructure is shown in Figure~\ref{fig:search_pipeline}.

As the first step we create distributed image indices from visualjoins using Hadoop.  Sharded with doc-ID, each machine contains indexes (and features) associated with a subset of the entire image collections.  Two types of indexes are used: the first is disk stored (and partially memory cached) \textit{token indices} with vector-quantized features (e.g. visual vocabulary) as key, and image doc-id hashes as posting lists. This is analogous to text based image retrieval system except text is replaced by visual tokens.  The second is memory cached features including both visual and meta-data such as image annotations and ``topic vectors" computed from the user-board-image graph. The first part is used for fast (but imprecise) lookup, and the second part is used for more accurate (but slower) ranking refinement.   

Each machine runs a leaf ranker, which first computes K-nearest-neighbor from the indices and then re-rank the top candidates by computing a score between the query image and each of the top candidate images based on additional metadata such as annotations. In some cases the leaf ranker skips the token index and directly retrieve the K-nearest-neighbor images from the feature tree index using variations of approximate KNN such as~\cite{binaryknn}.  A root ranker hosted on another machine will retrieve K top results from each of the leaf rankers, and them merge the results and return them to the users.  To handle new fingerprints generated with our real-time feature extractor, we have an online version of the visual search pipeline where a very similar process occurs. With the online version however, the given fingerprint is queried on pre-generated indices.

\section{Application 1: Related Pins}
\label{sec:app-related-pins}

\begin{figure}
\centering
\includegraphics[width=3 in]{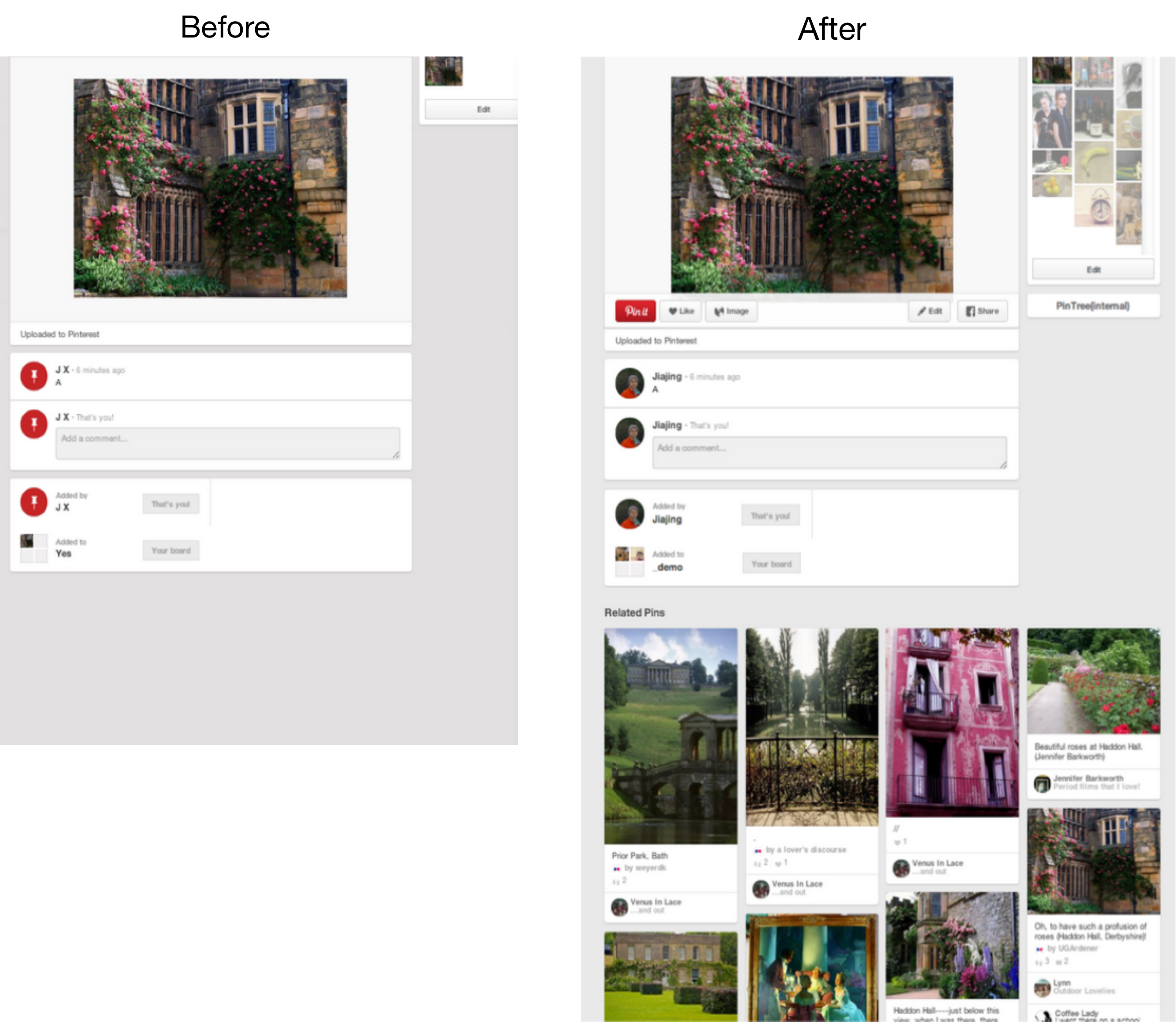}
\caption{Before and after incorporating Visual Related Pins}
\label{fig:beforeafter}
\end{figure}

One of the first applications of Pinterest's visual search pipeline was within a recommendations product called Related Pins, which recommends other images a user may be interested in when viewing a Pin. Traditionally, we have used a combination of user-curated image-to-board relationships and content-based signals to generate these recommendations. A problem with this approach, however, is that computing these recommendations is an offline process, and the image-to-board relationship must already have been curated, which may not be the case for our less popular Pins or newly created Pins.
As a result, 6\% of images at Pinterest have very few or no recommendations. For these images, we used the visual search pipeline described previously to generate \textit{Visual} Related Pins based on visual signals as shown in Figure~\ref{fig:beforeafter}. 

The first step of the Visual Related Pins product is to use the local token index built from all existing Pinterest images to detect if we have near duplicates to the query image. Specifically, given a query image, the system returns a set of images that are variations of the same image but altered through transformation such as resizing, cropping, rotation, translation, adding, deleting and altering minor parts of the visual contents. Since the resulting images look visually identical to the query image, their recommendations are most likely relevant to the query image. In most cases, however, we found that there are either no near duplicates detected or the near duplicates do not have enough recommendations. Thus, we focused most of our attention on retrieving visual search results generated from an index based on deep features.

\subsection*{Static Evaluation of Search Relevance}

Our initial Visual Related Pins experiment utilized features from the original and fine-tuned versions of the AlexNet model in its search infrastructure. However, recent successes with deeper CNN architectures for classification led us to investigate the performance of feature sets from a variety of CNN models. 

To conduct evaluation for visual search, we used the image annotations associated with the images as proxy for relevancy.  This approach is commonly used for offline evaluation of visual search systems~\cite{Muller:2001} in addition to human evaluation.   In this work, we used top text-queries associated each image as testing annotations.  We retrieve 3,000 images per query for 1000 queries using Pinterest Search, which yields a dataset with about 1.6 million unique images.  We label each image with the query that produced it. A visual search result is assumed to be relevant to a query image if the two images share a label.

\begin{table}[h]
\begin{center}
\label{visualsearchrel}
\caption{Relevance of visual search. }
\begin{tabular}{  l  c  c  r }
\hline
   Model &   p@5 &  p@10  &  latency   \\
\hline
AlexNet FC6	 &  0.051 & 0.040 & 193ms \\
Pinterest FC6	 & 0.234 & 0.210 & 234ms \\
GoogLeNet	 & 0.223 & 0.202 & 1207ms \\
VGG 16-layer	 & 0.302 & 0.269 & 642ms \\
\hline
\end{tabular}
\end{center}
\end{table}


Using this evaluation dataset, we computed the precision@k measure for several feature sets: the original AlexNet 6th layer fully-connected features (\textit{fc6}), the \textit{fc6} features of a fine-tuned AlexNet model trained with Pinterest product data, GoogLeNet (``loss3" layer output), and the \textit{fc6} features of the VGG 16-layer network~\cite{vggnet2014}.  We also examined combining the score from the aforementioned \textit{low-level features} with the score from the output vector of the classifier layer (the \textit{semantic features}). Table 1 shows p@5 and p@10 performance of these models using low level features for nearest neighbor search, along with the average latency of our visual search service (which includes feature extraction for the query image as well as retrieval). We observed a substantial gain in precision against our evaluation dataset when using the FC6 features of the VGG 16-layer model, with an acceptable latency for our applications.

\begin{figure}
	\centering
	\includegraphics[width=3.2in]{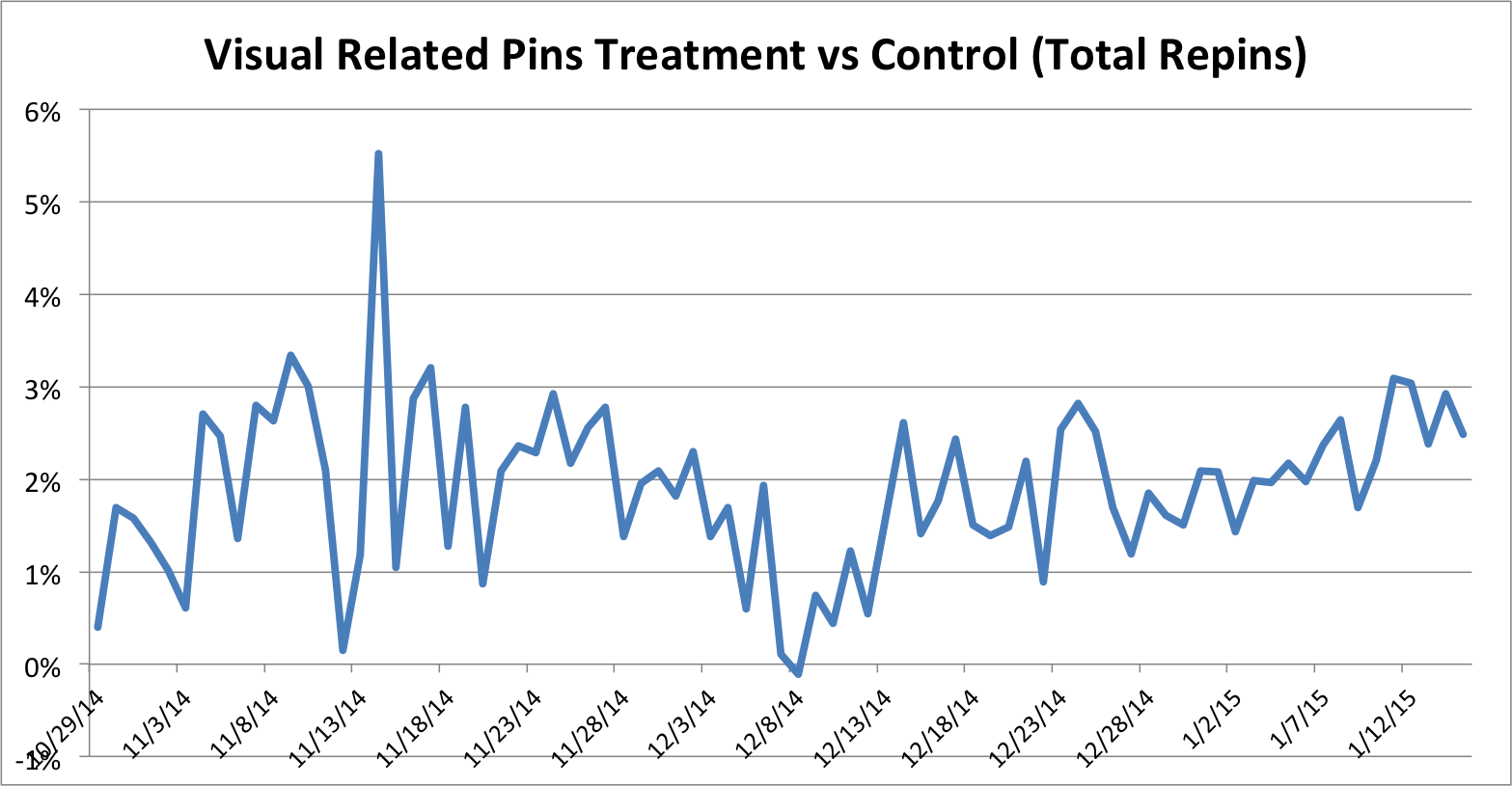}
	\caption{Visual Related Pins increases total Related Pins repins on Pinterest by 2\%.}
	\label{fig:visual_related_pins}
\end{figure}

\subsection*{Live Experiments}

For our experiment, we set up a system to detect new Pins with few recommendations, query our visual search system, and store their results in HBase to serve during Pin close-up.

%
%
One improvement we built on top of the visual search system for this experiment was adding a results metadata conformity threshold to allow greater precision at the expense of lower recall. This was important as we feared that delivering poor recommendations to a user would have lasting effects on that user's engagement with Pinterest. This was particularly concerning as our visual recommendations are served when viewing newly created Pins, a behavior that occurs often in newly joined users. As such we chose to lower the recall if it meant improving relevancy.

We launched the experiment initially to 10\% of Pinterest eligible live traffic. We considered a user to be eligible when they viewed a Pin close-up that did not have enough recommendations, and triggered a user into either a treatment group where we replaced the Related Pins section with visual search results, or a control group where we did not alter the experience. In this experiment, what we measured was the change in total repins in the Related Pins section where repinning is the action of a user adding an image to their collections. We chose to measure repins as it is one of our top line metrics and a standard metric for measuring engagement.

After running the experiment for three months, Visual Related Pins increased total repins in the Related Pins product by 2\% as shown in Figure~\ref{fig:visual_related_pins}.

\section{Application 2: Similar Looks}
\label{sec:app-similar-looks}

One of the most popular categories on Pinterest is women's fashion. However, a large percentage of pins in this category do not direct users to a shopping experience, and therefore aren't actionable.  There are two challenges towards making these Pins actionable: 1)
Many pins feature editorial shots such as ``street style" outfits, which often link to a website with little additional information on the items featured in the image;
2) Pin images often contain multiple objects (e.g. a woman walking down the street, with a leopard-print bag, black boots, sunglasses, torn jeans, etc.) A user looking at the Pin might be interested in learning more about the bag, while another user might want to buy their sunglasses.

User research revealed this to be a common user frustration, and our data indicated that users are much less likely to clickthrough to the external Website on women's fashion Pins, relative to other categories.

\begin{figure}
\centering
\includegraphics[width=3.2 in]{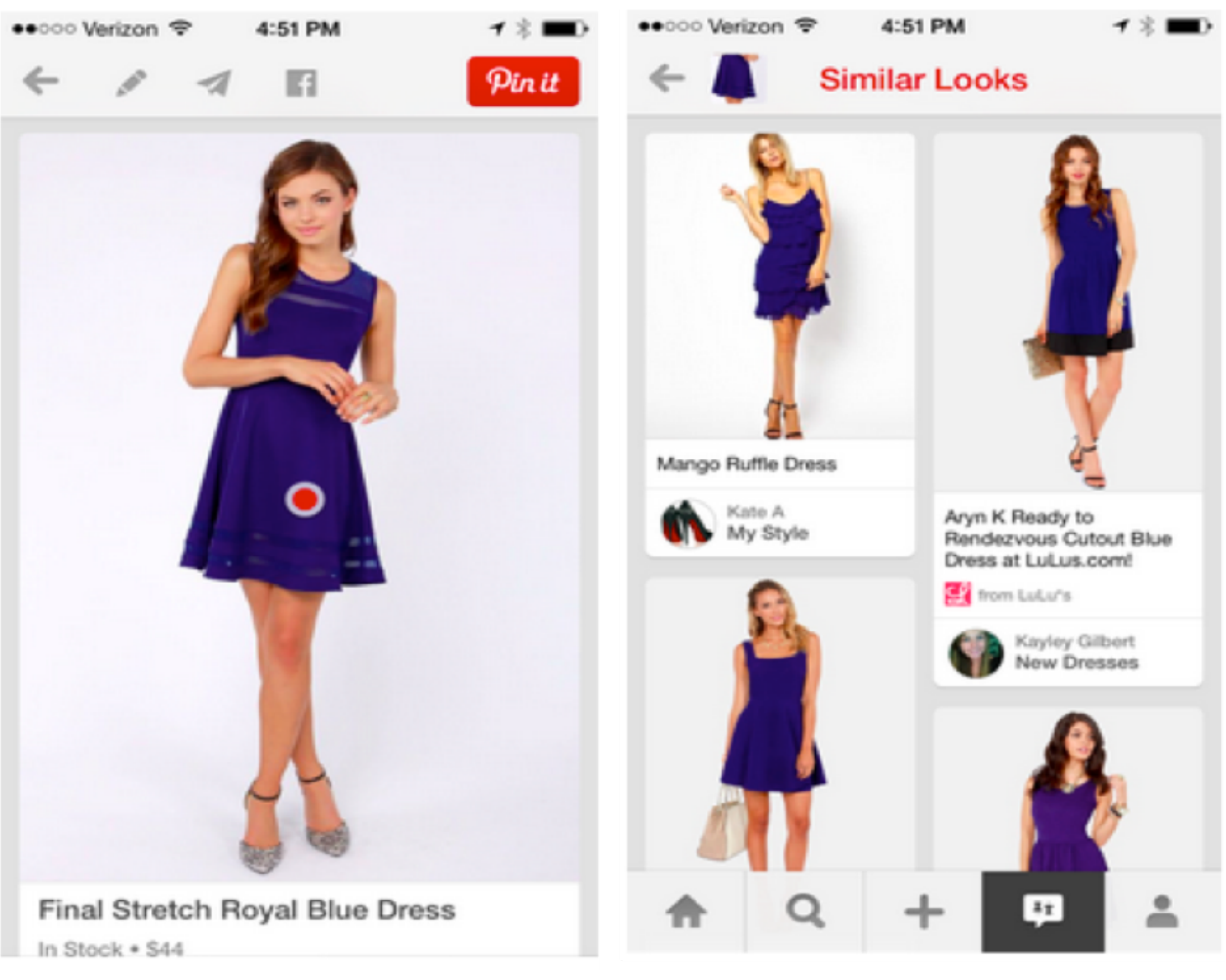}
\caption{Once a user clicks on the red dot, the system shows products that have a similar appearance to the query object.}
\label{fig:similarlooksmobile}
\end{figure}

To address this problem, we built a product called ``Similar Looks", which localized and classified fashion objects (Figure~\ref{fig:similarlooksmobile}). We use object recognition to detect products such as bags, shoes, pants, and watches in Pin images. From these objects, we extract visual and semantic features to generate product recommendations (``Similar Looks"). A user would discover the recommendations if there was a red dot on the object in the Pin (see Figure \ref{fig:similarlooks}). Clicking on the red dot loads a feed of Pins featuring visually similar objects (e.g. other visually similar blue dresses).

\subsection*{Related Work}

Applying visual search to ``soft goods" has been explored both within academia and industry.  Like.com, Google Shopping and Zappos (owned by Amazon) are a few well-known applications of computer vision to fashion recommendations.  Baidu and Alibaba also launched visual search systems recently solving similar problems. There is also a growing amount of research on vision-based fashion recommendations~\cite{fashion,streetshop,ebay}. Our approach demonstrates the feasibility of an object-based visual search system on tens of millions of Pinterest users and exposes an interactive search experience around these detected objects.

\subsection*{Static Evaluation of Object Localization}

The first step of evaluating our Similar Looks product was to investigate our object localization and detection capabilities. We chose to focus on fashion objects because of the aforementioned business need and because ``soft goods" tend to have distinctive visual shapes (e.g. shorts, bags, glasses).

We collected our evaluation dataset by randomly sampling a set of images from Pinterest's women's fashion category, and manually labeling 2,399 fashion objects in 9 categories (shoes, dress, glasses, bag, watch, pants, shorts, bikini, earnings) on the images by drawing a rectangular crop over the objects.  We observed that shoes, bags, dresses and pants were the four largest categories in our evaluation dataset. Shown in Table~\ref{table:objects} is the distribution of fashion objects as well as the detection accuracies from the text-based filter, image-based detection, and the combined approach (where text filters are applied prior to object detection). 

\label{sec:evaluationobject}
\begin{table}[h]
\caption{Object detection/classification accuracy (\%) }

\label{table:objects}
\begin{tabular}{  l l | l  l | l l | l l }
  &  & Text&   &  Img & & Both &   \\
 Objects & \# & TP &  FP  &  TP   & FP  & TP & FP   \\
\hline
shoe	 &873&79.8&6.0&41.8&3.1&34.4 &1.0 \\
dress	&383	&75.5	  &  6.2 &	58.8	& 12.3 & 47.0	& 2.0\\
glasses	&238	&75.2	& 18.8 & 63.0 & 0.4 &	50.0 &	0.2\\
bag	&468	&66.2	& 5.3	& 59.8 & 2.9	& 43.6	& 0.5\\
watch	&36	&55.6	& 6.0	& 66.7 &	0.5	& 41.7	& 0.0\\
pants	&253	&75.9	 & 2.0	& 60.9	& 2.2	& 48.2	& 0.1\\
shorts	&89	&73.0	& 10.1 & 44.9 &1.2	& 31.5	& 0.2\\
bikini	&32	&71.9	& 1.0	& 31.3 &0.2	& 28.1	& 0.0\\
earrings	&27	&81.5	& 4.7	& 18.5 &0.0	& 18.5	& 0.0\\
\hline
Average & & 72.7 & 6.7 & 49.5 & 2.5 & 38.1 & 0.5\\
\hline
\end{tabular}
\end{table}

As previously described, the text-based approach applies manually crafted rules (e.g. regular expressions) to the Pinterest meta-data associated with images (which we treat as weak labels).  For example, an image annotated with ``spring fashion, \textit{tote} with flowers" will be classified as ``bag," and is considered as a positive sample if the image contains a ``bag" object box label.  For image-based evaluation, we compute the intersection between the predicted object bounding box and the labeled object bounding box of the same type, and count an intersection to union ratio of 0.3 or greater as a positive match.


Table~\ref{table:objects} demonstrates that neither text annotation filters nor object localization alone were sufficient for our detection task due to their relatively high false positive rates at 6.7\% and 2.5\% respectively. Not surprisingly, combining two approaches significantly decreased our false positive rate to less than 1\%.

Specifically, we saw that for classes like ``glasses" text annotations were insufficient and image-based classification excelled (due to a distinctive visual shape of glasses). For other classes, such as ``dress", this situation was reversed (the false positive rate for our dress detector was high, 12.3\%, due to occlusion and high variance in style for that class, and adding a text-filter dramatically improved results). Aside from reducing the number of images we needed to fingerprint with our object classifiers, for several object classes (shoe, bag, pants), we observed that text-prefiltering was crucial to achieve an acceptable false positive rate (1\% or less).

\subsection*{Live Experiments} 

Our system identified over 80 million ``clickable'' objects from a subset of Pinterest images. A clickable red dot is placed upon the detected object.  Once the user clicks on the dot, our visual search system retrieves a collection of Pins most visually similar to the object. We launched the system to a small percentage of Pinterest live traffic and collected user engagement metrics such as CTR for a period of one month. Specifically, we looked at the clickthrough rate of the dot, the clickthrough rate on our visual search results, and also compared engagement on Similar Look results with the existing Related Pin recommendations.

\begin{figure}
\centering
\includegraphics[width=3 in]{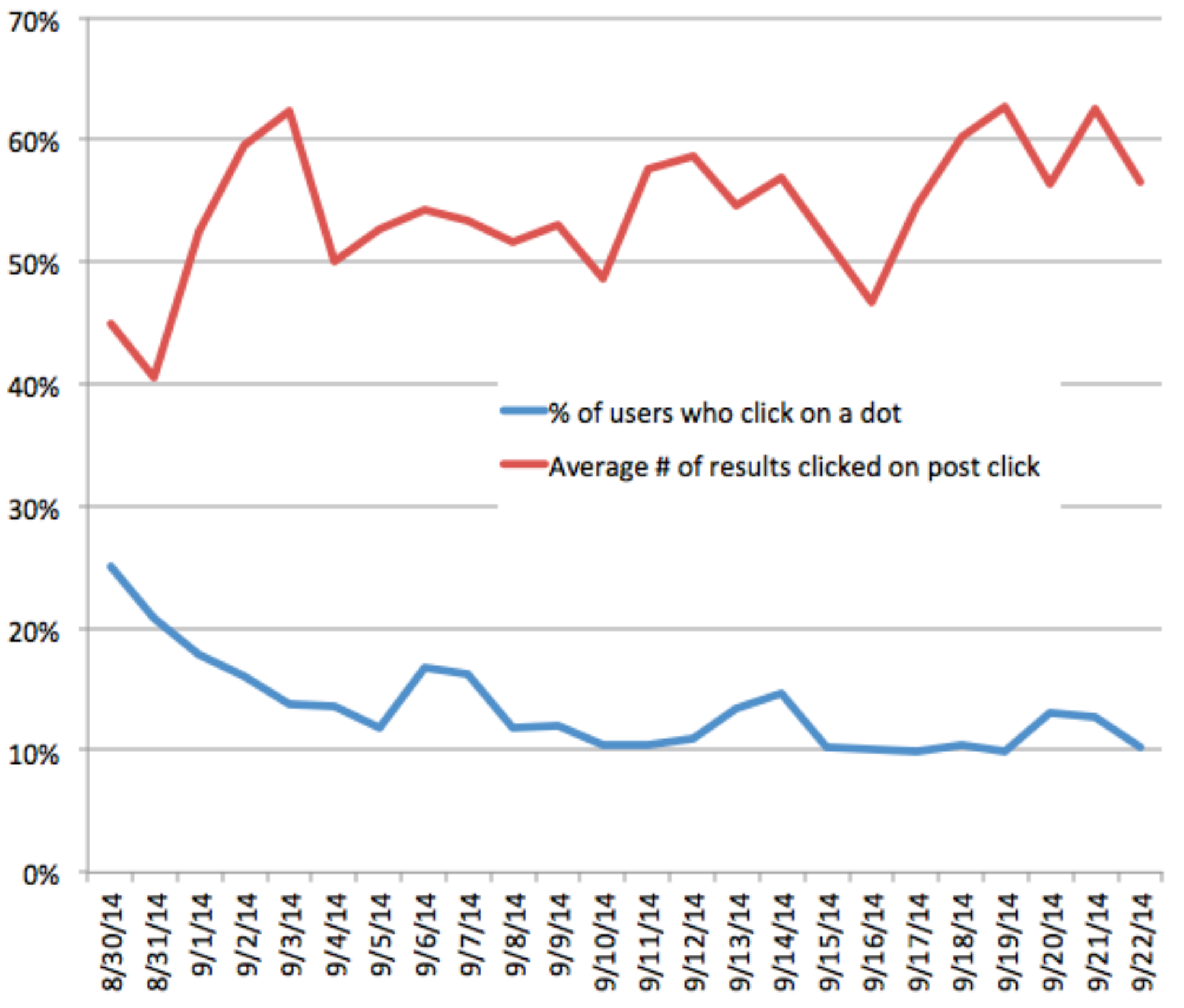}
\caption{Engagement rates for Similar Looks experiment}
\label{fig:similarlooksresults}
\end{figure}

As shown in Figure~\ref{fig:similarlooksresults}, an average of 12\% of users who viewed a pin with a dot clicked on a dot in a given day. Those users went on to click on an average 0.55 Similar Look results. Although this data was encouraging, when we compared engagement with all related content on the pin close-up (summing both engagement with Related Pins and Similar Look results for the treatment group, and just related pin engagement for the control), Similar Looks actually hurt overall engagement on the pin close-up by 4\%. After the novelty effort wore off, we saw gradual decrease in CTR on the red dots which stabilizes at around 10\%. 

To test the relevance of our Similar Looks results independently of the bias resulting from the introduction of a new user behavior (learning to click on the ``object dots"), we designed an experiment to blend Similar Looks results directly into the existing Related Pins product (for Pins containing detected objects). This gave us a way to directly measure if users found our visually similar recommendations relevant, compared to our non-visual recommendations.  On pins where we detected an object, this experiment increased overall engagement (repins and close-ups) in Related Pins by 5\%. Although we set an initial static blending ratio for this experiment (one visually similar result to three production results), this ratio adjusts in response to user click data.



\begin{figure}
\centering
\includegraphics[width=3.5 in]{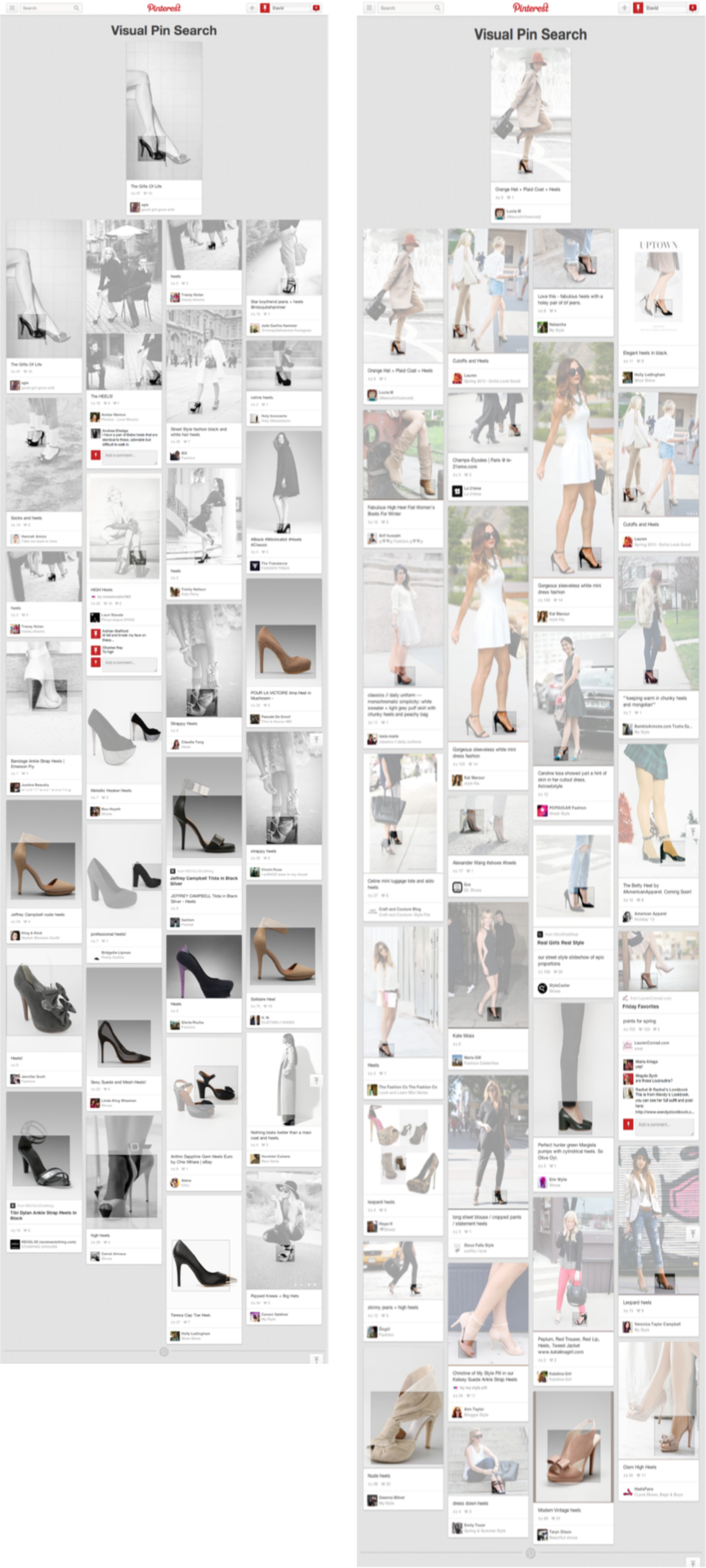}
\caption{Examples of object search results for shoes. Boundaries of detected objects are automatically highlighted.  The top image is the query image.}
\end{figure}

\begin{figure*}
\centering
\includegraphics[width=7 in]{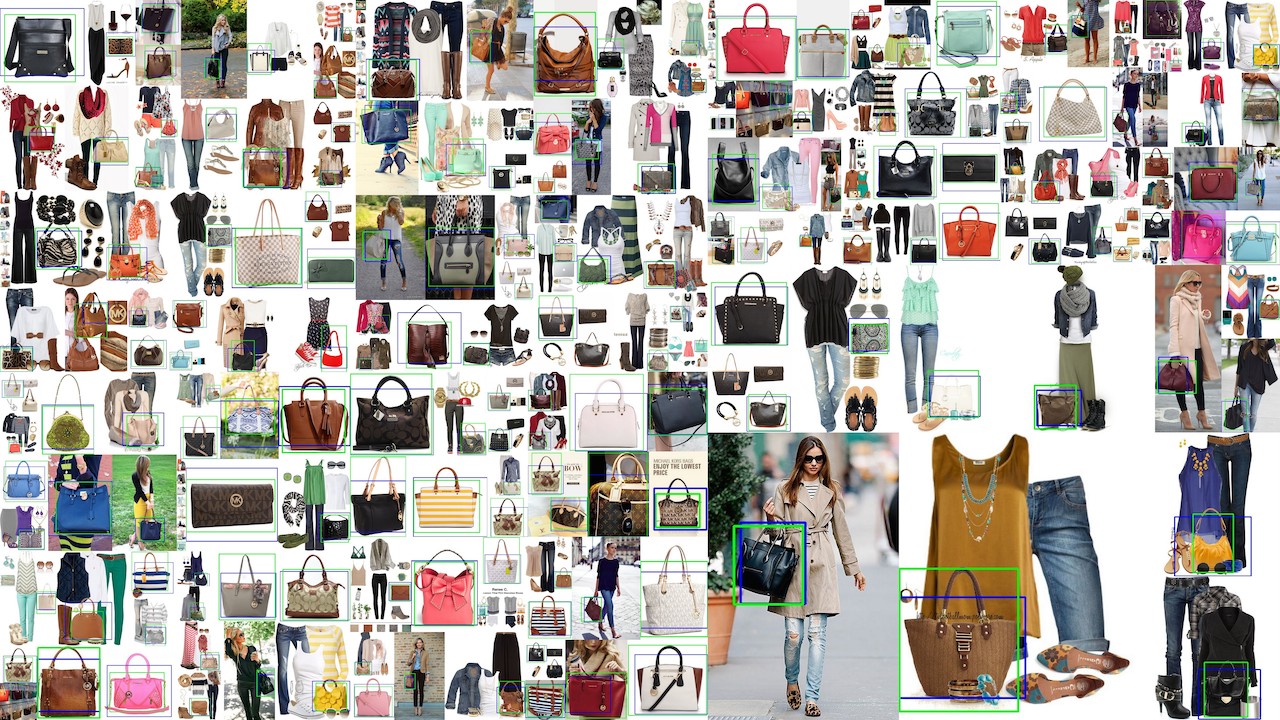}
\caption{Samples of object detection and localization results for bags. [Green: ground truth, blue: detected objects.] }
\end{figure*}

\begin{figure*}
\centering
\includegraphics[width=7 in]{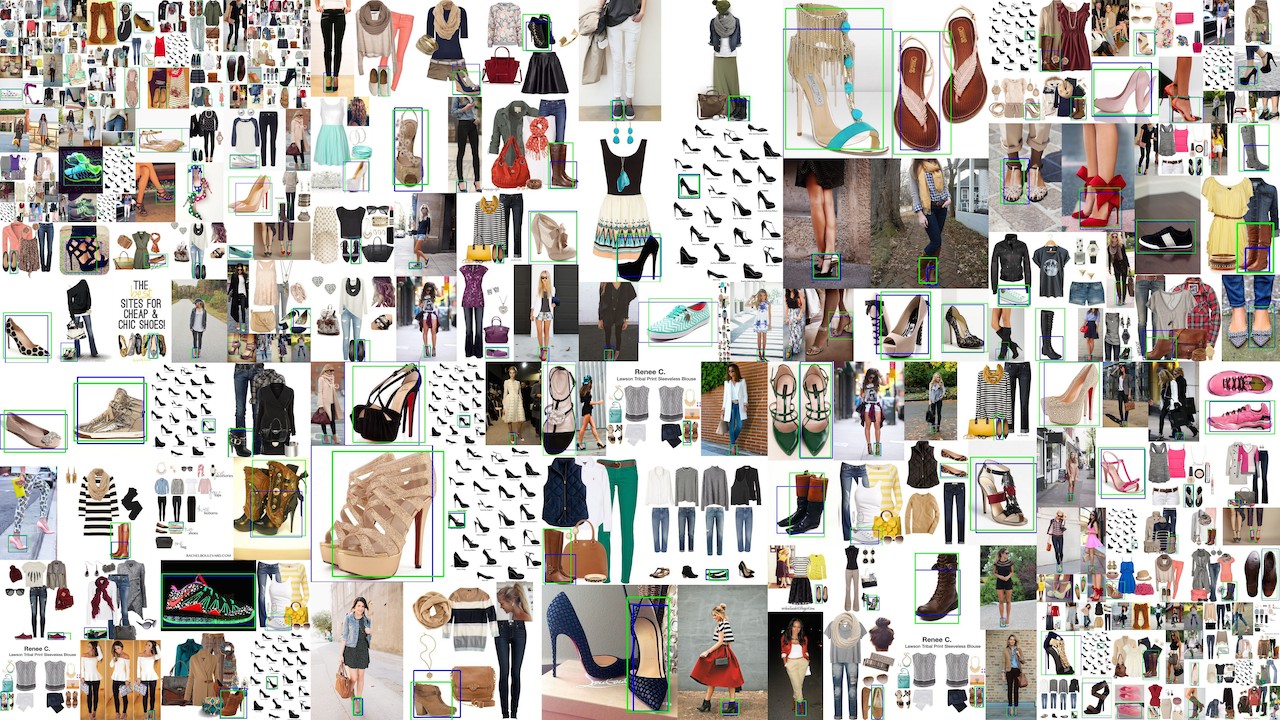}
\caption{Samples of object detection and localization results for shoes.}
\end{figure*}

\section{Conclusion and Future Work}

We demonstrate that, with the availability of distributed computational platforms such as Amazon Web Services and open-source tools, it is possible for a handful of engineers or an academic lab to build a large-scale visual search system using a combination of non-proprietary tools.  This paper presented our end-to-end visual search pipeline, including incremental feature updating and two-step object detection and localization method that improves search accuracy and reduces development and deployment costs.  Our live product experiments demonstrate that visual search features can increase user engagement.

We plan to further improve our system in the following areas.
First, we are interested in investigating the performance and efficiency of CNN based object detection methods in the context of live visual search systems. Second, we are interested in leveraging Pinterest ``curation graph" to enhance visual search relevance. Lastly, we want to experiment with alternative interactive interfaces for visual search.


\bibliographystyle{abbrv}
\bibliography{citation}
%





\end{document}